\documentclass[conference]{IEEEtran}
\IEEEoverridecommandlockouts
\usepackage{cite}
\usepackage{amsmath,amssymb,amsfonts}
\usepackage{bm}
\usepackage{algorithm}
\usepackage{algpseudocode}
\usepackage{graphicx}
\usepackage{textcomp}
\usepackage{xcolor}
\usepackage{multirow}
\usepackage{booktabs}
\usepackage{subcaption}
\usepackage[shortlabels,inline]{enumitem}

\newcommand{\ours}[0]{\textsc{Valce}}
\newcommand{\gcn}[0]{\textsc{Gcn}}
\newcommand{\sage}[0]{\textsc{GraphSAGE}}
\newcommand{\gat}[0]{\textsc{Gat}}
\newcommand{\mpnn}[0]{\textsc{Mpnn}}
\newcommand{\congnet}[0]{\textsc{CongestionNet}}
\newcommand{\ptp}[0]{\textsc{Pix2pix}}
\newcommand{\lhnn}[0]{\textsc{Lhnn}}
\newcommand{\cgnn}[0]{\textsc{CircuitGnn}}

\def\BibTeX{{\rm B\kern-.05em{\sc i\kern-.025em b}\kern-.08em
    T\kern-.1667em\lower.7ex\hbox{E}\kern-.125emX}}
\begin{document}

\title{Variational Label-Correlation Enhancement for Congestion Prediction}

\author{\IEEEauthorblockN{
Biao Liu\IEEEauthorrefmark{1},
Congyu Qiao\IEEEauthorrefmark{1},
Ning Xu\IEEEauthorrefmark{1},
Xin Geng\IEEEauthorrefmark{1},
Ziran Zhu\IEEEauthorrefmark{2}, and
Jun Yang\IEEEauthorrefmark{2}}
\IEEEauthorblockA{\IEEEauthorrefmark{1}School of Computer Science and Engineering,\\ Southeast University, Nanjing 211189, China\\ Email: liubiao01@seu.edu.cn, qiaocy@seu.edu.cn, xning@seu.edu.cn, xgeng@seu.edu.cn}
\IEEEauthorblockA{\IEEEauthorrefmark{2}School of Integrated Circuits,\\ Southeast University, Nanjing 211189, China\\ Email: zrzhu@seu.edu.cn, dragon@seu.edu.cn}
}


\maketitle

\begin{abstract}


The physical design process of large-scale designs is a time-consuming task, often requiring hours to days to complete, with routing being the most critical and complex step. 
As the the complexity of Integrated Circuits (ICs) increases, there is an increased demand for accurate routing quality prediction.  Accurate congestion prediction aids in identifying design flaws early on, thereby accelerating circuit design and conserving resources.
Despite the advancements in current congestion prediction methodologies, an essential aspect that has been largely overlooked is the spatial label-correlation between different grids in congestion prediction. The spatial label-correlation is a fundamental characteristic of circuit design, where the congestion status of a grid is not isolated but inherently influenced by the conditions of its neighboring grids. In order to fully exploit the inherent spatial label-correlation between neighboring grids, we propose a novel approach, {\ours}, i.e., VAriational Label-Correlation Enhancement for Congestion Prediction, which considers the local label-correlation in the congestion map, associating the estimated congestion value of each grid with a local label-correlation weight influenced by its surrounding grids. {\ours} leverages variational inference techniques to estimate this weight, thereby enhancing the regression model's performance by incorporating spatial dependencies. Experiment results validate the superior effectiveness of {\ours} on the public available \texttt{ISPD2011} and \texttt{DAC2012} benchmarks using the superblue circuit line.

\end{abstract}


\section{Introduction}

Integrated Circuits (ICs) are fundamental components of contemporary electronic goods, including computers, smartphones, and vehicles. The design process of these circuits involves the use of Electronic Design Automation (EDA) tools \cite{karn2000eda}, which facilitate the different stages of development, most notably the logic synthesis and placement stages. Given the ever-increasing scale and intricacy of circuits, optimizing the efficiency and accuracy of EDA tools has emerged as a crucial challenge, prompting researchers to leverage deep learning techniques to enhance the circuit design process  \cite{bustany2015ispd, chen2020pros, ma2020understanding}.

As part of these efforts to improve the IC design process using deep learning, a key area of focus is congestion prediction. Congestion, in the context of IC design, refers to areas in the chip layout where there is a high density of components and interconnections. High congestion can lead to numerous design issues, such as increased delay and power consumption, and can even cause the circuit to fail. Therefore, accurate and early prediction of congestion is vital, as it allows design flaws to be identified and corrected before the physical chip is manufactured, saving both time and resources.

In the field of EDA, a variety of methods have been developed to predict congestion. These methods can be broadly divided into topological and geometrical approaches, each with their unique focus and techniques. Topological methods primarily concentrate on the logical relationships within circuit designs, often utilizing Graph Neural Networks (GNNs) \cite{zhang2019heterogeneous} to capture the intricate interactions among cells for accurate congestion prediction \cite{8920342, ghose2021generalizable, veli2018graph}. Geometrical methods, on the other hand, emphasize the spatial information of circuit designs, leveraging diverse techniques ranging from image-based representations \cite{zhou2019congestion} to lattice networks \cite{wang2022lhnn}. Additionally, some innovative methods have attempted to merge topological and geometrical information into a unified data structure for improved prediction \cite{yang2022versatile}.

Despite the advancements in current congestion prediction methodologies, an essential aspect that has been largely overlooked is the spatial label-correlation between different grid cells in congestion prediction. This correlation is based on the principle that the congestion status of a grid cell is not isolated but is inherently influenced by the conditions of its neighboring cells. That is, a grid with high congestion typically implies a higher likelihood of its neighboring grids also being congested due to the spatial continuity of circuit elements and interconnections. Unfortunately, current methods have not adequately addressed this spatial dependency, leading to a gap in the predictive capabilities of existing approaches.

In this paper, we consider the label-correlation in the congestion map and assume that the estimated congestion value of the each grid in the circuit design is associated with a label-correlation weight constituted by the positive real number of its surrounding grids, representing the degree to each surrounding grid influencing the congestion value of the center grid. Generally, a grid with a higher congestion value tends to indicate higher congestion values in its surrounding grids. This kind of label-correlation is because a higher congestion value often implies more pins, which typically connect to nearby circuit components and can result in increased congestion values for neighboring grids. Hence, label-correlation is an essential geometrical relationship contained in the congestion map and the corresponding label-correlation weight is worth being estimated to further regularize the risk estimator for the regression model.

Motivated by the above consideration, we deal with the problem of congestion prediction from two aspects. First, we enhance the spatial label-correlation of the estimated congestion map by estimating a label-correlation weight employing the variational inference technique \cite{2013Auto}. Second, we iteratively estimate the label-correlation weight and train the regression model with a regularized risk estimator involved with label-correlation. The proposed method named {\ours}, i.e., \textit{VAriational Label-Correlation Enhancement for congestion prediction}, estimates the label-correlation weights via inferring the variational posterior density parameterized by an inference model with the deduced evidence lower bound, and trains the regression model with a risk estimator by leveraging the ground-truth congestion value as well as the label-correlation weights. In summary, our contributions are:
\begin{itemize}[topsep=0ex,leftmargin=*,parsep=1pt,itemsep=1pt]
	\item  We for the first time consider the label-correlation in the congestion map, i.e., the estimated congestion value of each grid is associated with that of its surrounding grids, which is the intuitive geometrical relationship modeled by a latent label-correlation and worth being estimated for regression model training.
	\item  We train the regression model with a proposed regularized risk estimator by leveraging the label-correlation. The posterior density of the latent label-correlation weight is inferred via taking on the approximate Gamma density parameterized by an inference model and deduce the evidence lower bound for optimization.
	\item  Experiment results validate the superior effectiveness of {\ours} on the public available ISPD2011 and DAC2012 benchmarks using the superblue circuit line.
\end{itemize}

\section{Related Work}

The related work section is structured around two key methodological categories: the topological methods and the geometrical methods, which are primarily applied during the logic synthesis stage the placement stage respectively.

The topological methods in EDA primarily concentrate on the logical relationships between cells and nets, often transforming circuit designs into graph representations consisting of vertices and edges \cite{ren2022graph}. These methods provide a structured approach to understanding and analyzing the interconnections and dependencies within the circuit design.
One such method, {\congnet} \cite{8920342}, as well as solutions proposed in \cite{ghose2021generalizable}, establish connections between cell pairs that are linked through nets. These approaches utilize popular GNNs, such as the Graph Attention Network ({\gat}) \cite{veli2018graph}, to generate cell representations that can be used for congestion prediction. By employing GNNs, these methods are able to effectively capture the complex relationships and interactions among cells, resulting in more accurate congestion prediction.

The geometrical methods in EDA primarily focus on the spatial information of circuit designs, aiming to utilize the layout and positioning of components to provide insights into congestion prediction. There are several approaches to incorporate this information, with varying degrees of complexity and effectiveness.

A popular approach employed by geometrical methods involves dividing the circuit into small rectangles, referred to as grids, and converting the design into a representation akin to RGB channels, where the grids function as pixels \cite{yu2019painting, xie2018routenet}. By adopting this image-based representation, the methods can leverage powerful image processing techniques to analyze and predict congestion in the circuit layout. Subsequently, image translation techniques (e.g., {\ptp} \cite{zhou2019congestion}) are employed to generate new images with the red channels populated, indirectly addressing the congestion prediction problem through image analysis.

Additionally, {\lhnn} \cite{wang2022lhnn} adopts a different approach, converting circuits into lattice networks \cite{you2017deep} rather than images. In this representation, each grid serves as an internal node in the network, while each net, as an external node, is connected to the grid it covers geometrically. This approach facilitates a more direct representation of the spatial relationships within the circuit layout. Furthermore, the {\cgnn} \cite{yang2022versatile} presents a versatile graph neural network that designs a heterogeneous graph, called the Circuit Graph, to integrate both topological and geometrical information into a unified data structure. By employing a message-passing and fusion approach named {\cgnn}, the method learns a powerful representation for congestion prediction, effectively combining the strengths of both topological and geometrical methods.

These geometrical methods, tailored for the placement stage, showcase the potential of incorporating spatial information to enhance congestion prediction in EDA. By leveraging various representations and techniques, they provide a diverse set of tools to tackle the complex challenges posed by congestion in circuit designs.

\section{Methodology}

\begin{figure*}[t]
	\centering
	\includegraphics[width=\textwidth]{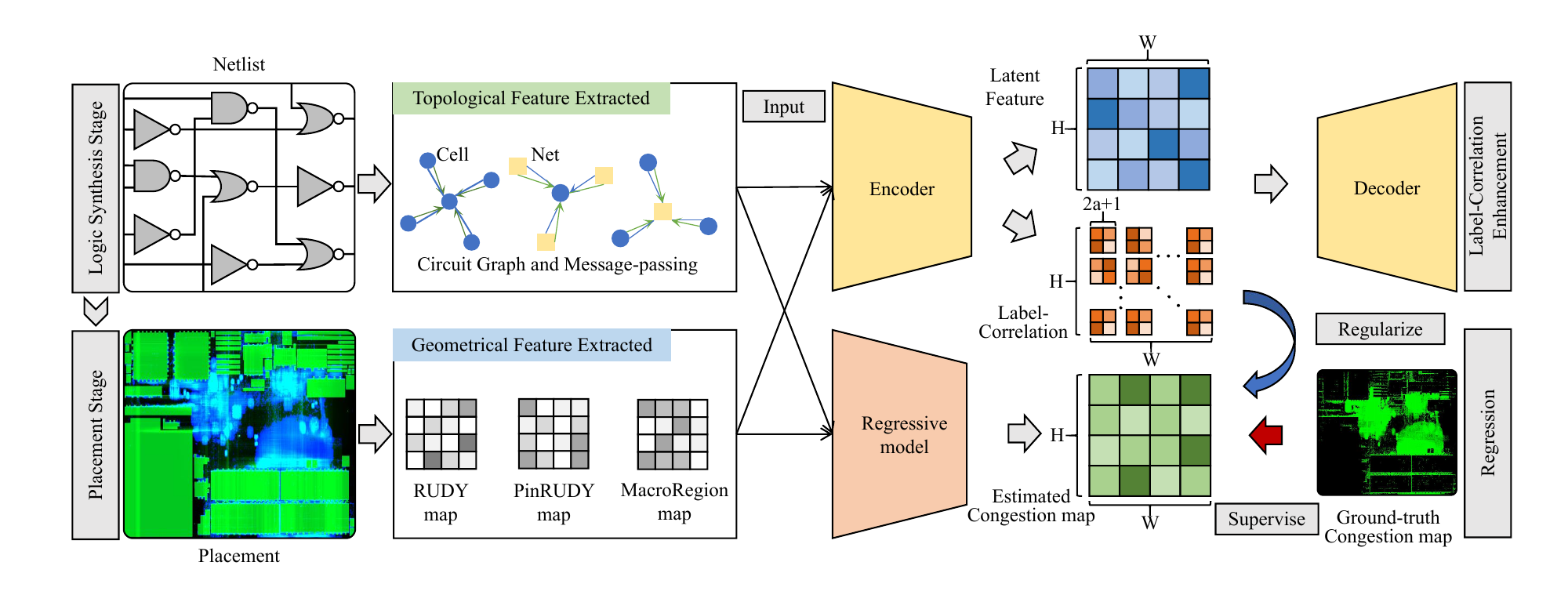}
	\caption{An illustration of the working flow for our congestion prediction approach.}
	\label{fig:working_flow}
\end{figure*}

\subsection{Notations}

Let $\mathcal{X}\subset \mathbb{R}^q$ denote $q$-dimensional feature space and $\mathcal{Y}\subset \mathbb{R}_{+}^{H\times W}$ denote the congestion map space with $H\times W$ positive real values, which represents the degree of congestion in each grid. The training dataset is denoted by $\mathcal{D}=\{(\mathbf{X}_i,\mathbf{Y}_i) \vert 1\leq i\leq n\}$ where each feature of the circuit design $\mathbf{X}_i \in \mathcal{X}$ and its associated congestion map $\mathbf{Y}_i = [\bm y_i^1, \bm y_i^2, ..., \bm y_i^H]\in \mathcal{Y}$ with $\bm{y}_i^j=[y_i^{j,1}, y_i^{j,2}, ..., y_i^{j,W}]$. We formulate the task of congestion prediction in EDA as a regression problem. Our goal is to learn a regression model $f$: $\mathcal{X}\mapsto \mathcal{Y}$, which could predict the congestion map on the unseen feature of the circuit design as accurately as prossible.

\subsection{The overall framework}


In this paper, we further consider label-correlation, i.e., the geometric relationship at the label-level between each grid and its surrounding grids, to improve the prediction accuracy. Intuitively, the larger the congestion value of a grid is, the larger the congestion values of its surrounding grids may be. This is because the larger congestion value of a grid usually means more pins, and these pins tend to connect surrounding circuit components, which may lead to the larger congestion value of its surrounding grids. Hence, By enhancing the label-correlation, the performance of the regression model could be further improved via the following regularized risk:
\begin{equation}\label{loss:risk}
	\begin{aligned}
		\hat{R}(f) = \frac{1}{n}\sum_{i=1}^{n}(\mathcal{L}_{\text{sup}}(f(\mathbf{X}_i), \mathbf{Y}_i) + \lambda \mathcal{L}_{\text{reg}}(f(\mathbf{X}_i), \mathbf{M}_i)),
	\end{aligned}
\end{equation}
where we model the label-correlation for each example as a latent variable $\mathbf{M}_i$ and employ the Variational Inference (VI) technique to estimate it in the next subsection, $\mathcal{L}_{\text{sup}}$ is a conventional loss function (such as mean squared error and mean absolute error) for regression, which measures how well a model estimates a given real-valued label, and the multiplicative factor $\lambda$ is used to balance the contribution of these two loss terms. 

\begin{figure*}[t]
	\centering
	\includegraphics[width=\textwidth]{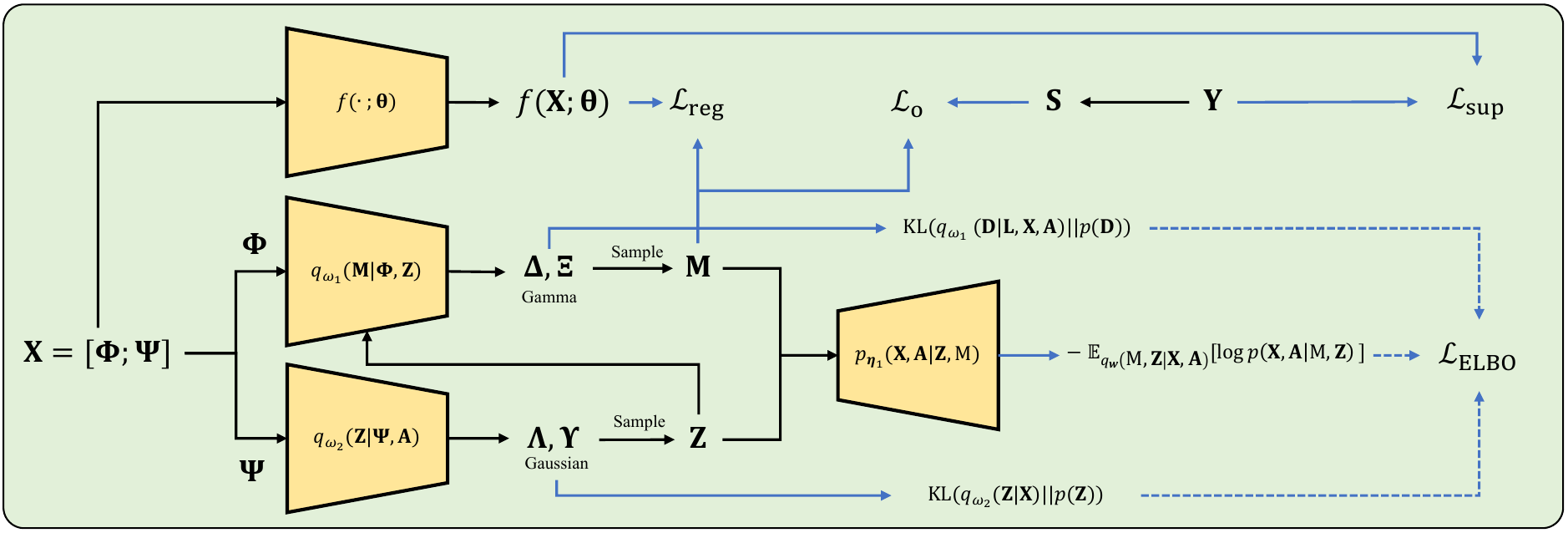}
	\caption{An illustration of the optimization flow of {\ours}. }
	\label{fig:variational_label_correlation_working_flow}
\end{figure*}

We iteratively estimate the label-correlation weight and train the regression model with a regularized risk estimator that incorporates label-correlation. The working and optimization flow of our approach is illustrated in Figure \ref{fig:working_flow} and \ref{fig:variational_label_correlation_working_flow}.

\subsection{Variational Label-Correlation Enhancement}

Given the feature $\mathbf{X}_i$ and its estimated congestion map $f(\mathbf{X}_i)$, we consider the neighborhood of the estimated value $f^{j,k}(\mathbf{X}_i)$, which is denoted by $\mathcal{N}_{a}(f^{j,k}(\mathbf{X}_i))$, as follows:
\begin{equation}
	\begin{aligned}
		&\mathcal{N}_{a}(f^{j,k}(\mathbf{X}_i)) = \{f^{j+h,k+w}(\mathbf{X}_i) | h,w\in \mathbb{N}, \\
		&-a \leq h \leq a, -a \leq w \leq a, 1 \leq j \leq H, \\
        &1 \leq k \leq W\},
	\end{aligned}
\end{equation}
where $a$ is a integer which reflects the size of the neighborhood, and we let $f^{j+h,k+w}=f^{j,k}$ if $j+h\notin [1,H]$ or $v+w \notin [1, W]$ for convenience.

Let $\mathbf{M}_i=[\mathbf{M}_i^{1,1}, ...,\mathbf{M}_i^{1,W}; ...; \mathbf{M}_i^{H,1}, ...,\mathbf{M}_i^{H, W}]$ be the latent tensor to control the degree of label-correlation, where $\mathbf{M}_i^{j, k} = [m_{i,j,k}^{1,1}, ..., m_{i,j,k}^{1,2a+1};...;m_{i,j,k}^{2a+1,1}, ..., m_{i,j,k}^{2a+1,2a+1}]$ is the label-correlation weight for each $\mathbf{M}_i$ at the location $(j,k)$. Then, we formulate the label-correlation loss function of each $f^{j,k}(\mathbf{X}_i)$ and its neighborhood $\mathcal{N}_{a}(f^{j,k}(\mathbf{X}_i))$ as follows:

\begin{equation}\label{loss:lwc}
	\begin{aligned}
		&\ell_{\text{lc}}(f^{j,k}(\mathbf{X}_i), \mathbf{M}_i^{j, k})=
		\sum_{h=1}^{2a+1}\sum_{w=1}^{2a+1} m_{i,j,k}^{h, w} \cdot \\
  &\vert \vert f^{j,k}(\mathbf{X}_i) - f^{r,v}(\mathbf{X}_i) \vert \vert, 
	\end{aligned}
\end{equation}
where $r=j+h-a-1$ and $v=k+w-a-1$.

Based on Eq.(\ref{loss:lwc}), $\mathcal{L}_{\text{reg}}(f(\mathbf{X}_i), \mathbf{M}_i))$ can be calculated as follows:
\begin{equation}\label{loss:reg}
	\begin{aligned}
		\mathcal{L}_{\text{reg}}(f(\mathbf{X}_i), \mathbf{M}_i))=\sum_{j=1}^{H}\sum_{k=1}^{W} \ell_{\text{lc}}(f^{j,k}(\mathbf{X}_i), \mathbf{M}_i^{j, k}).
	\end{aligned}
\end{equation}

Next, we will employ the variational inference technique to estimate the label-correlation weight tensor $\mathbf{M}_i$ for each circuit design.

\begin{algorithm}[t] 
	\caption{ {\ours} Algorithm} 
	\label{alg:Framework} 
	\begin{algorithmic}[1] 
		\Require 
		The training set $\mathcal{D}=\{(\mathbf{X}_i,\mathbf{Y}_i) \vert 1\leq i\leq n\}$, epoch $T$ and iteration $I$;
		\State Initialize the regressive model $\bm{\theta}$,  the reference model $\bm{w}=[\bm{w}_1,\bm{w}_2]$ and observation model $\bm{\eta}$;
		\State Extract the geometrical features $\{\mathbf{\Phi}_1, \mathbf{\Phi}_2, ..., \mathbf{\Phi}_n\}$ and the topological features $\{\mathbf{\Psi}_1, \mathbf{\Psi}_2, ..., \mathbf{\Psi}_n\}$, and calculate the adjacency matrix $\{\mathbf{A}_1, \mathbf{A}_2, ..., \mathbf{A}_n\}$;
		
		\For {$t=1,\ldots, T$}

		\State Shuffle training set $\mathcal{D}$ into $I$ mini-batches;
		\For {$k=1,\ldots, I$}
		\State Obtain the label-correlation weight tensor $\mathbf{M}_i$ for each example $\mathbf{X}_i$ by Eq. (\ref{eq:gamma_posterior}); 
		\State Update  $\bm{\theta}$, $\bm{w}$ and $\bm{\eta}$ by forward computation and back-propagation by fusing Eq. (\ref{loss:risk}) and Eq. (\ref{loss:vi}); 
		\EndFor
		\EndFor
		\Ensure The regressive model $\bm{\theta}$.
	\end{algorithmic} 
\end{algorithm}

We split each feature $\mathbf{X}_i$ into the geometrical and topological features,i.e.,$\mathbf{X}_i=[\mathbf{\Phi}_i;\mathbf{\Psi}_i]$, where  $\mathbf{\Phi}_i\in \mathbb{R}^{H\times W \times a}$ denotes the geometrical feature and $\mathbf{\Psi}_i \in \mathbb{R}^{C \times b}$ denotes topological features where $C$ denotes the number of cells.  To predict each weight value $m_{i,j,k}^{h,w} \in R^{+}$ of $\mathbf{M}_i$, we treat $m_{i,j,k}^{h,w}$ as a latent variable, whose prior density $p(m_{i,j,k}^{h,w})$ is a Gamma density with the minor value $\hat{\alpha}_{i,j,k}^{h,w}$ and $\hat{\beta}_{i,j,k}^{h,w}$ as its parameters, i.e., $p(m_{i,j,k}^{h,w})=\text{Gamma}(m_{i,j,k}^{h,w} | \hat{\alpha}_{i,j,k}^{h,w}, \hat{\beta}_{i,j,k}^{h,w})$. Then prior density $p(\textbf{M}_i)$  can be represent as the product of each Gamma:
\begin{equation}\label{eq:gamma_prior}
	\begin{aligned}
		p(\textbf{M}_i) = \prod_{j=1}^{H}\prod_{k=1}^{W}\prod_{h=1}^{2a+1}\prod_{w=1}^{2a+1} \text{Gamma}(m_{i,j,k}^{h,w} | \hat{\alpha}_{i,j,k}^{h,w}, \hat{\beta}_{i,j,k}^{h,w})
	\end{aligned}
\end{equation}

We also assume that there exists a latent feature matrix $\mathbf{Z}_i = [\bm z_i^1, \bm z_i^2, ..., \bm z_i^C]^T$ dependent on the topological information, which the topological feature is generated from. Here, we let $\bm z_i^j=[z_i^{j,1}, z_i^{j,2}, ..., z_i^{j,b}]$ and each density $p(z_i^{j,k})$ is a 
standard Gaussian with the mean $\hat{\mu}_i^{j,k}=0$ and standard deviation $\hat{\sigma}_i^{j,k}=1$, i.e, $p(z_i^{j,k})=\text{Gaussian}(z_i^{j,k} | \hat{\mu}_i^{j,k}, \hat{\sigma}_i^{j,k})$. Then we let the prior density $p(\textbf{Z}_i)$ be the product of each Gaussian:

\begin{equation}\label{eq:gaussian_prior}
	\begin{aligned}
		p(\textbf{Z}_i) = \prod_{j=1}^{C}\prod_{k=1}^{b} \text{Gaussian}(z_i^{j,k} | \hat{\mu}_i^{j,k}, \hat{\sigma}_i^{j,k})
	\end{aligned}
\end{equation}

Let each geometrical feature $\mathbf{\Phi}_i$ and topological feature $\mathbf{\Psi}_i$, the adjacency matrix of each circuit design $\mathbf{A}_i$ be observed, where $\mathbf{A}_i \in \mathbb{R}^{C\times C}$ with each value $a_i^{j,k}=1$ if cell $j$ is connected to cell $k$, otherwise $a_i^{j,k}=0$. {\ours} aims to infer the posterior density $p(\mathbf{M}_i|\mathbf{\Phi}_i, \mathbf{Z}_i)$ and $p(\mathbf{Z}_i|\mathbf{\Psi}_i, \mathbf{A}_i)$.

Due to the computational complexity associated with obtaining the exact posterior density $p(\mathbf{M}_i|\mathbf{\Phi}_i, \mathbf{Z}_i)$ and $p(\mathbf{Z}_i|\mathbf{\Psi}_i, \mathbf{A}_i)$, the utilization of fixed-form density $q(\mathbf{M}_i|\mathbf{\Phi}_i, \mathbf{Z}_i)$ and $q(\mathbf{Z}_i|\mathbf{\Psi}_i, \mathbf{A}_i)$ serves as a practical means of approximating the true posterior distribution. We let the approximate posterior $q(\mathbf{M}_i|\mathbf{\Phi}_i, \mathbf{Z}_i)$ be the product of each Gamma parameterized by $\alpha_i^{j,k}$ and $\beta_i^{j,k}$:

\begin{equation}\label{eq:gamma_posterior}
	\begin{aligned}
		q_{\bm \omega_1}(\mathbf{M}_i|\mathbf{\Phi}_i, \mathbf{Z}_i)&= \prod_{j=1}^{H}\prod_{k=1}^{W}\prod_{h=1}^{2a+1}\prod_{w=1}^{2a+1} \\
        &\text{Gamma}(m_{i,j,k}^{h,w} | \alpha_{i,j,k}^{h,w}, \beta_{i,j,k}^{h,w}).
	\end{aligned}
\end{equation}

Here, let $\mathbf{\Delta}_i=[\mathbf{\Delta}_i^{1,1}, ..., \mathbf{\Delta}_i^{1,W};...;\mathbf{\Delta}_i^{H,1}, ..., \mathbf{\Delta}_i^{H,W}]$ and $\mathbf{\Xi}_i =[\mathbf{\Xi}_i^{1,1}, ..., \mathbf{\Xi}_i^{1,W};...;\mathbf{\Xi}_i^{H,1}, ..., \mathbf{\Xi}_i^{H,W}]$ with the matrix $\bm \Delta_i^{j,k}=[\alpha_{i,j,k}^{1,1}, ..., \alpha_{i,j,k}^{1,2a+1};...;\alpha_{i,j,k}^{2a+1,1}, ..., \alpha_{i,j,k}^{2a+1,2a+1}]$ and $\bm \Xi_i^{j,k}=[\beta_{i,j,k}^{1,1}, ..., \beta_{i,j,k}^{1,2a+1};...;\beta_{i,j,k}^{2a+1,1}, ..., \beta_{i,j,k}^{2a+1,2a+1}]$ are the outputs of the inference model parameterized by $\bm \omega_1$, given the geometrical feature $\mathbf{\Phi}_i$ and the latent topological feature $\mathbf{Z}_i$.

Also, we let the approximate posterior $q(\mathbf{Z}_i|\mathbf{\Psi}_i, \mathbf{A}_i)$ be the product of each Gaussian parameterized by $\mu_i^{j,k}$ and $\sigma_i^{j,k}$:

\begin{equation} \label{eq:gaussian_posterior}
	\begin{aligned}
		q_{\bm \omega_2}(\mathbf{Z}_i|\mathbf{\Psi}_i, \mathbf{A}_i)=\prod_{j=1}^{C}\prod_{k=1}^{b}\text{Gaussian}(z_i^{j,k} | \mu_i^{j,k}, \sigma_i^{j,k}).
	\end{aligned}
\end{equation}

Here, the parameters $\mathbf{\Lambda}_i=[\bm \mu_i^1, \bm \mu_i^2, ..., \bm \mu_i^r]$ and $\mathbf{\Upsilon}_i =[\bm \sigma_i^1, \bm \sigma_i^2, ..., \bm \sigma_i^r]$ with the vector $\bm \mu_i^j=[\mu_i^{j,1}, \mu_i^{j,2}, ..., \mu_i^{j,c}]^T$ and $\bm \sigma_i^j=[\sigma_i^{j,1}, \sigma_i^{j,2}, ..., \sigma_i^{j,c}]^T$ are the outputs of the inference model parameterized by $\bm \omega_2$, given the input topological feature $\mathbf{\Psi}_i$ and the adjacency matrix $\mathbf{A}_i$.

Adopting the variational inference framework, the derivation of the evidence lower bound (ELBO) for the model's marginal likelihood guarantees the optimization of $q_{\bm \omega_1}(\mathbf{M}_i|\mathbf{\Phi}_i, \mathbf{Z}_i)$ and $q_{\bm \omega_2}(\mathbf{Z}_i |\mathbf{\Psi}_i, \mathbf{A}_i)$ to closely approximate $p(\mathbf{M}_i|\mathbf{\Phi}_i, \mathbf{Z}_i)$ and $p(\mathbf{Z}_i|\mathbf{\Psi}_i, \mathbf{A}_i)$, respectively:

\begin{equation}\label{loss:elbo}
	\begin{aligned}
		&\mathcal{L}_{\text{ELBO}} = \mathbb{E}_{q_{\bm \omega_1, \omega_2}(\mathbf{M},\mathbf{Z} | \mathbf{\Phi}, \mathbf{\Psi}, \mathbf{A})}[\log p_{\bm \eta}(\mathbf{X}, \mathbf{A} | \mathbf{M}, \mathbf{Z})] \\ 
		&- \text{KL}(q_{\bm{w}_1}(\mathbf{M} | \mathbf{\Phi},\mathbf{Z})|| p(\mathbf{M})) \\
        &- \text{KL}(q_{\bm{w}_2}(\mathbf{Z} | \mathbf{\Psi},\mathbf{A})|| p(\mathbf{Z}))
	\end{aligned}
\end{equation}

As the first part of Eq.(\ref{loss:elbo}) is intractable, we employ the implicit reparameterization trick \cite{NEURIPS2018_92c8c96e} to approximate it by Monte Carlo (MC) estimation. Note that we can use only one MC sample in Eq. (\ref{loss:elbo}) during the training process as suggested in \cite{2013Auto, xu2020variational}. Then the first part of Eq.(\ref{loss:elbo}) is tractable:
\begin{equation} \label{loss:rec}
    \begin{aligned}
        &\mathbb{E}_{q_{\bm \omega_1, \omega_2}(\mathbf{M},\mathbf{Z} | \mathbf{\Phi}, \mathbf{\Psi}, \mathbf{A})}[\log p_{\bm \eta}(\mathbf{X}, \mathbf{A} | \mathbf{M}, \mathbf{Z})] = \\
		&\frac{1}{n} \sum_{i=1}^n \Vert \widehat{\bm\Phi}_i - \bm\Phi_i \Vert_F^2 + \Vert \widehat{\bm\Psi}_i - \bm\Psi_i \Vert_F^2 \\
		& \qquad + \Vert \mathcal{S}(\mathbf Z_i\mathbf Z_i^T) - \mathbf A_i \Vert_F^2,
    \end{aligned}
\end{equation}
where $ \widehat{\bm\Phi}_i $ and $ \widehat{\bm\Psi}_i $ are reconstructed geometrical and topological features with two layer Multi-Layer Perceptrons (MLPs) \cite{riedmiller1994advanced} parameterized by $\bm \eta$, respectively. $\mathcal{S}(\cdot)$ is the sigmoid function.

According to Eq.(\ref{eq:gamma_prior}) and Eq.(\ref{eq:gamma_posterior}), the first KL divergence in Eq.(\ref{loss:elbo}) can be analytically calculated as follows:
\begin{equation}\label{loss:kl_gamma}
	\begin{aligned}
		&\text{KL}(q_{\bm{w}_1}(\mathbf{M} | \mathbf{\Phi},\mathbf{Z})|| p(\mathbf{M})) = \sum_{i=1}^{n} \sum_{j=1}^{H} \sum_{k=1}^{W} \sum_{h=1}^{2a+1} \sum_{w=1}^{2a+1} \\
		&\hat{\alpha}_{i,j,k}^{h,w}\log \frac{\beta_{i,j,k}^{h,w}}{\hat{\beta}_{i,j,k}^{h,w}} - \log 
		\frac{\Gamma(\alpha_{i,j,k}^{h,w})}{\Gamma(\hat{\alpha}_{i,j,k}^{h,w})} + (\alpha_{i,j,k}^{h,w}-\hat{\alpha}_{i,j,k}^{h,w}) \cdot \\
		& \psi(\alpha_{i,j,k}^{h,w}) - (\beta_{i,j,k}^{h,w}-\hat{\beta}_{i,j,k}^{h,w})\frac{\alpha_{i,j,k}^{h,w}}{\beta_{i,j,k}^{h,w}},
	\end{aligned}
\end{equation}
where $\Gamma(\cdot)$ and $\psi(\cdot)$ are Gamma function and Digamma function, respectively.

According to Eq.(\ref{eq:gaussian_prior}) and Eq.(\ref{eq:gaussian_posterior}), the second KL divergence in Eq.(\ref{loss:elbo}) can be analytically calculated as follows:

\begin{equation}\label{loss:kl_gaussian}
	\begin{aligned}
		&\text{KL}(q_{\bm{w}_2}(\mathbf{Z} | \mathbf{\Psi},\mathbf{A})|| p(\mathbf{Z})) = \sum_{i=1}^{n} \sum_{j=1}^{C} \sum_{k=1}^{b} \big(\log \frac{\hat{\mu}_{i}^{j,k}}{\mu_i^{j,k}} \\ 
		&+ \frac{(\sigma_i^{j,k})^2 + (\mu_i^{j,k}-\hat{\mu}_i^{j,k})^2}{2(\hat{\sigma}_i^{j,k})^2} - \frac{1}{2}\big)
	\end{aligned}
\end{equation}

Furthermore, we introduce the compatibility loss, a regularization term that ensures that the label-correlation weight will not depart from the local similarity in the ground-truth congestion map $\mathbf{Y}$:

\begin{equation}\label{loss:compatibility}
	\begin{aligned}
		\mathcal{L}_o = \sum_{i=1}^{n}\vert\vert \mathbf{M}_i - \mathbf{S}_i \vert\vert,
	\end{aligned}
\end{equation}
where $\mathbf{S}_i=[\mathbf{S}_i^{1,1}, ...,\mathbf{S}_i^{1,W}; ...; \mathbf{S}_i^{H,1}, ...,\mathbf{S}_i^{H, W}]$ denotes the local similarity tensor in the ground-truth congestion map $\mathbf{Y}_i$ with  $\mathbf{S}_i^{j, k} = [s_{i,j,k}^{1,1}, ..., s_{i,j,k}^{1,2a+1};...;s_{i,j,k}^{2a+1,1}, ..., s_{i,j,k}^{2a+1,2a+1}]$ as the local similarity matrix. Each local similarity value will be calculated as follows:

\begin{equation}\label{def:similarity}
	\begin{aligned}
		s_{i,j,k}^{h,w} = e^{-\frac{||y_i^{j,k} - y_i^{h,w}||}{2\sigma^2}}
	\end{aligned}
\end{equation}

Now we can easily get the optimization objective of variational label-correlation enhancement as follows
\begin{equation}\label{loss:vi}
	\begin{aligned}
		\mathcal{L}_{\text{VI}} = \tau \mathcal{L}_{o} - \mathcal{L}_{\text{ELBO}},
	\end{aligned}
\end{equation}
 
where $\tau$ is a hyper-parameter. The estimated congestion map $\mathbf{M}$ is sampled from $q(\mathbf{M} \vert \mathbf{\Phi}, \mathbf{Z})$.

\subsection{Practical implementations}

The overall procedure of our framework is present in Algorithm \ref{alg:Framework}. To train the regressive model, we minimize the empirical risk in Eq.(\ref{loss:risk}), where the regressive model $f$ parameterized by $\bm \theta$ and we adopt the average value $\mathbf{M}_i$ sampled by $\mathbf{M}_i \sim q_{\bm{w}_1}(\mathbf{M}_i | \mathbf{\Phi},\mathbf{Z}_i)$. Meanwhile, the label-correlation weight $\mathbf{M}_i$ will updated as the models in variational inference are updated via minimizing Eq.(\ref{loss:vi}).

\textbf{Feature Extracted. } On the one hand, for the geometrical feature $\mathbf{\Phi}$, we accord with the experimental setup detailed in \cite{liu2021global}. each geometrical feature composing the input $M\times N\times 3$ feature map from the cell placement solution, including the $M\times N$ RUDY \cite{4211973} map, the $M\times N$ PinRUDY map and the $M\times N$ MacroRegion map. On the other hand, for the topological feature $\mathbf{\Psi}$, we accord with the experimental setup detailed in \cite{yang2022versatile}. The proposed methodology involves transforming the Circuit Design into a Circuit Graph, which is then subjected to MLPs to initialize the features of cells, nets, topo-edges, and geom-edges into hidden representations. Subsequently, these hidden representations undergo a sequence of circuit message-passing layers, resulting in deeper representations of the cells and nets. Ultimately, the output cell and net representations are leveraged for our framework.

\textbf{Logic Synthesis Stage. } Our framework aims at the placement stage, where the geometric feature and topological feature are both provided. Hence, when it comes to the logic synthesis stage, where the geometric feature is not included, some adjustments need to be made accordingly. Eq.(\ref{loss:vi}) will degenerate as follows:
\begin{equation}\label{loss:vi_2}
	\begin{aligned}
		\mathcal{L}'_{\text{VI}} = - \mathcal{L}'_{\text{ELBO}},
	\end{aligned}
\end{equation}
where
\begin{equation}
	\begin{aligned}
		\mathcal{L}'_{\text{ELBO}} = \mathbb{E}_{q_{\bm \omega_2}(\mathbf{Z} | \mathbf{\Psi}, \mathbf{A})}[\log p(\mathbf{\Psi}, \mathbf{A} | \mathbf{Z})] \\
		- \text{KL}(q_{\bm{w}_2}(\mathbf{Z} | \mathbf{\Psi},\mathbf{A})|| p(\mathbf{Z})).
	\end{aligned}
\end{equation}

Similar to Eq.(\ref{loss:rec}), the first part could be calculated as:
\begin{equation} \label{loss:rec2}
    \begin{aligned}
        &\mathbb{E}_{q_{\bm \omega_1, \omega_2}(\mathbf{M},\mathbf{Z} | \mathbf{\Phi}, \mathbf{\Psi}, \mathbf{A})}[\log p(\mathbf{\Psi}, \mathbf{A} | \mathbf{Z})] = \\
		&\frac{1}{n} \sum_{i=1}^n \Vert \widehat{\bm\Psi}_i - \bm\Psi_i \Vert_F^2 + \Vert \mathcal{S}(\mathbf Z_i\mathbf Z_i^T) - \mathbf A_i \Vert_F^2,
    \end{aligned}
\end{equation}
and the second part could be calculated as the same as Eq.(\ref{loss:kl_gaussian}).

When the latent feature $\textbf{Z}_i$ is captured by minimizing Eq.(\ref{loss:vi_2}), we adopt the average value $\textbf{Z}_i$ sampled by $\textbf{Z}_i \sim q_{\bm{w}_2}(\mathbf{Z} | \mathbf{\Psi}_i,\mathbf{A})$ as the latent topological feature value of the $i$-th circuit design, which will be input into the model $f$ to perform regression:
\begin{equation}\label{loss:risk2}
	\begin{aligned}
		\hat{R}(f) = \frac{1}{n}\sum_{i=1}^{n}\mathcal{L}_{\text{sup}}(f(\mathbf{Z}_i), \mathbf{Y}_i).
	\end{aligned}
\end{equation}

\section{Experiments}
\begin{table*}[t]
	\caption{Congestion prediction result in logic synthesis stage of \texttt{ISPD2011}.}
	\label{tab:logic_ispd}
	\centering
	\normalsize
		\begin{tabular}{ccccccc}
		\toprule                                                                                                                   
		\multirow{2}{*}{Baseline} & \multicolumn{3}{c}{Cell-level}                                    & \multicolumn{3}{c}{Grid-level}                                     \\ 
		\cmidrule(r){2-4} \cmidrule(l){5-7}
								& pearson             & spearman             & kendall              & pearson              & spearman             & kendall              \\ \midrule
		{\gcn} & 0.949 & 0.438 & 0.398 & 0.026 & 0.787 & 0.680 \\
		{\sage} & 0.948 & 0.566 & 0.522 & 0.080 & 0.798 & 0.713 \\
		{\gat} & 0.956 & 0.550 & 0.507 & 0.119 & 0.796 & 0.706 \\
		{\congnet} & \textbf{0.974} & 0.567 & 0.523 & \textbf{0.325} & 0.717 & 0.577 \\
		{\mpnn} & 0.952 & 0.429 & 0.341 & 0.234 & 0.142 & 0.067 \\
		{\cgnn} & 0.950 & 0.446 & 0.356 & 0.257 & 0.260 & 0.151 \\
		{\ours}(w/o. geom.) & \textbf{0.974} & \textbf{0.577} & \textbf{0.543} & 0.119 & \textbf{0.822} & \textbf{0.734} \\
		\bottomrule
		\end{tabular}
\end{table*}

\begin{table*}[t]
	\caption{Congestion prediction result in logic synthesis stage of \texttt{DAC2012}.}
	\label{tab:logic_dac}
	\centering
	\normalsize
		\begin{tabular}{ccccccc}
		\toprule                                                                                                                   
		\multirow{2}{*}{Baseline} & \multicolumn{3}{c}{Cell-level}                                    & \multicolumn{3}{c}{Grid-level}                                     \\ 
		\cmidrule(r){2-4} \cmidrule(l){5-7}
		& pearson             & spearman             & kendall              & pearson              & spearman             & kendall              \\ \midrule
		{\gcn} & 0.420 & 0.105 & 0.079 & -0.028 & 0.082 & 0.062 \\
		{\sage} & 0.464 & 0.118 & 0.094 & 0.026 & 0.183 & 0.130 \\
		{\gat} & 0.761 & 0.036 & 0.026 & 0.035 & -0.071 & -0.039 \\
		{\congnet} & 0.772 & 0.163 & 0.126 & 0.085 & 0.204 & 0.145 \\
		{\mpnn} & \textbf{0.806} & 0.294 & 0.225 & 0.255 & 0.476 & 0.343 \\
		{\cgnn} & 0.804 & 0.290 & 0.221 & 0.264 & 0.486 & 0.349 \\
		{\ours}(w/o. geom.) & 0.796 & \textbf{0.342} & \textbf{0.312} & \textbf{0.269} & \textbf{0.612} & \textbf{0.523} \\
		\bottomrule
		\end{tabular}
\end{table*}

\subsection{Datasets}
In the experiments, two widely-used datasets, \texttt{ISPD2011} \cite{viswanathan2011ispd} and \texttt{DAC2012} \cite{viswanathan2012dac}, are employed to evaluate the performance of {\ours}. The \texttt{ISPD2011} dataset consists of multiple designs, among which designs 1/2/4/5/10/12 are selected for training, design 15 is utilized for validation, and design 19 serves as the testing set. Similarly, the \texttt{DAC2012} dataset is divided into training, validation, and testing sets, with designs 2/3/6/7/9/11/12/14 designated for training, design 16 for validation, and design 19 for testing.

DREAMPlace \cite{lin2019dreamplace}, an open-source placement engine, is adopted to place cells and initialize the raw features of cells, nets, and grids. In order to generate the congestion targets on the grids, NCTU-GR 2.0 \cite{liu2013nctu}, a popular global router, is employed. The congestion target for each cell is set according to the value of the grid in which it is located.

Two distinct stages of the design process are considered for congestion prediction: the logic synthesis stage and the placement stage. During the logic synthesis stage, the topology of the circuits and the geometry-insensitive features are utilized for prediction. In the placement stage, additional geometry-sensitive features are incorporated to further enhance the prediction accuracy.

\subsection{Correlation metrics for evaluation}

\begin{table*}[t]
	\caption{Congestion prediction result in placement stage of \texttt{ISPD2011}.}
	\label{tab:place_ispd}
	\centering
	\normalsize
		\begin{tabular}{ccccccc}
		\toprule                                                                                                                   
		\multirow{2}{*}{Baseline} & \multicolumn{3}{c}{Cell-level}                                    & \multicolumn{3}{c}{Grid-level}                                     \\ 
		\cmidrule(r){2-4} \cmidrule(l){5-7}
		& pearson             & spearman             & kendall              & pearson              & spearman             & kendall              \\ \midrule
		{\gat} (w. geom.) & 0.959 & 0.568 & 0.524 & 0.112 & 0.803 & 0.717 \\
		{\ptp} & - & - & - & \textbf{0.419} & 0.399 & 0.318 \\
		{\lhnn} & - & - & - & -0.030 & 0.019 & 0.016 \\
		{\cgnn} (w/o. topo.) & 0.969 & 0.573 & 0.539 & 0.134 & 0.813 & 0.726 \\
		{\cgnn} & 0.965 & 0.571 & 0.538 & 0.182 & 0.809 & 0.722 \\
		{\ours} & \textbf{0.974} & \textbf{0.579} & \textbf{0.545} & 0.352 & \textbf{0.823} & \textbf{0.737} \\
		\bottomrule
		\end{tabular}
\end{table*}

\begin{table*}[t]
	\caption{Congestion prediction result in placement stage of \texttt{DAC2012}.}
	\label{tab:place_dac}
	\centering
	\normalsize
		\begin{tabular}{ccccccc}
		\toprule                                                                                                                   
		\multirow{2}{*}{Baseline} & \multicolumn{3}{c}{Cell-level}                                    & \multicolumn{3}{c}{Grid-level}                                     \\ 
		\cmidrule(r){2-4} \cmidrule(l){5-7}
		& pearson             & spearman             & kendall              & pearson              & spearman             & kendall              \\ \midrule
		{\gat} (w. geom.) & 0.704 & 0.135 & 0.107 & 0.029 & 0.189 & 0.134 \\
		{\ptp} & - & - & - & 0.337 & 0.304 & 0.238 \\
		{\lhnn} & - & - & - & 0.246 & 0.167 & 0.135 \\
		{\cgnn} (w/o. topo.) & 0.189 & 0.249 & 0.187 & 0.091 & 0.298 & 0.207 \\
		{\cgnn} & 0.223 & 0.237 & 0.178 & 0.128 & 0.321 & 0.225 \\
		{\ours} & \textbf{0.830} & \textbf{0.547} & \textbf{0.428} & \textbf{0.481} & \textbf{0.644} & \textbf{0.458} \\
		\bottomrule
		\end{tabular}
\end{table*}

Following the evaluation methodology employed in \cite{ghose2021generalizable}, the performance of {\ours} is assessed by comparing the predicted results with the ground-truth. Three widely-used correlation metrics, namely \textit{Pearson} \cite{cohen2009pearson}, \textit{Spearman} \cite{myers2004spearman}, and \textit{Kendall} \cite{abdi2007kendall}, are adopted to measure the strength and direction of the association between the predicted and ground-truth values on both cell level and grid level.

\begin{enumerate}[1)]

\item \textit{Pearson Correlation $ r $ }: The Pearson correlation coefficient is a parametric measure that quantifies the linear relationship between two continuous variables. Given a pair of random variables $ (X, Y) $, the Pearson correlation coefficient is defined as the covariance of $ X $ and $ Y $ divided by the product of their standard deviations. It is calculated using the following formula:
\begin{equation}
	\begin{aligned}
		r = \frac{\sum_{i=1}^{n}(x_i - \bar{x})(y_i - \bar{y})}{\sqrt{\sum_{i=1}^{n}(x_i - \bar{x})^2}\sqrt{\sum_{i=1}^{n}(y_i - \bar{y})^2}},
	\end{aligned}
\end{equation}
where $ x_i $ and $ y_i $  represent the individual data points of variables $ X $  and $ Y $, $ \bar{x} $ and $ \bar{y} $ are their respective means, and n is the total number of data points.

\item \textit{Spearman Correlation $ \rho $ }: The Spearman correlation coefficient is a non-parametric measure that evaluates the strength and direction of the monotonic relationship between two variables by considering their ranks. It is computed using the following formula:
\begin{equation}
	\begin{aligned}
		\rho = \frac{\sum_{i=1}^{n}(r_{x_i} - \bar{r}_x)(r_{y_i} - \bar{r}_y)}{\sqrt{\sum_{i=1}^{n}(r_{x_i} - \bar{r}_x)^2}\sqrt{\sum_{i=1}^{n}(r_{y_i} - \bar{r}_y)^2}},
	\end{aligned}
\end{equation}
where $ r_{x_i} $ and $ r_{y_i} $ represent the rank values of data points $ x_i $ and $ y_i $, $ \bar{r}_x $ and $ \bar{r}_y $ are the mean rank values of variables $ X $  and $ Y $ , and n is the total number of data points.

\item \textit{Kendall Correlation $ \tau $ }: The Kendall correlation coefficient is another non-parametric measure that assesses the degree of similarity between the orderings of two variables. It is calculated by:
\begin{equation}
	\begin{aligned}
		\tau=\frac{2}{n(n-1)} \sum_{i<j} \operatorname{sgn}\left(x_i-x_j\right) \operatorname{sgn}\left(y_i-y_j\right),
	\end{aligned}
\end{equation}
where $ \operatorname{sgn} $ is the sign function. This is the only metric used in \cite{8920342}, We add the other two metrics for more complete evaluation as especially Pearson can
capture raw values which Spearman and Kendall cannot.

\end{enumerate}

All of the correlation metrics range from $ -1 $  to $ 1 $ , where $ -1 $  indicates a perfect negative correlation, $ 1 $  signifies a perfect positive correlation, and $ 0 $  suggests no correlation. A higher absolute value of the three correlation metrics implies a stronger association between the predicted results and ground-truth.

\subsection{Baselines}

{\ours} is compared with other existing approaches to evaluate its performance and effectiveness. The comparison is conducted across two distinct stages of the design process, the logic synthesis stage and the placement stage. 

For the logic synthesis stage, {\ours} is benchmarked against alternative congestion prediction techniques that specifically focus on the topology of the circuits. Typical graph representation models {\gcn} \cite{kipf2017semisupervised}, {\sage} \cite{hamilton2017inductive}, {\gat} \cite{veli2018graph} and {\mpnn} \cite{gilmer2017neural}. Additionally, we also carry out results on EDA-customized meachine learning model including: 
\begin{enumerate}
	\item {\congnet} \cite{8920342} a multi-layer graph attentive architecture designed to predict circuit congestion,
	\item {\cgnn} \cite{yang2022versatile} a versatile graph neural network that designs a heterogeneous graph, Circuit Graph, to integrate topological and geometrical information
	into a unified data structure. In the logic synthesis stage, only the topology of the circuits is used as the features.
\end{enumerate}

\begin{figure*}[t]
	\centering
	\begin{subfigure}{0.25\linewidth}
	\centering
	\scalebox{1}[-1]{\includegraphics[width=\linewidth]{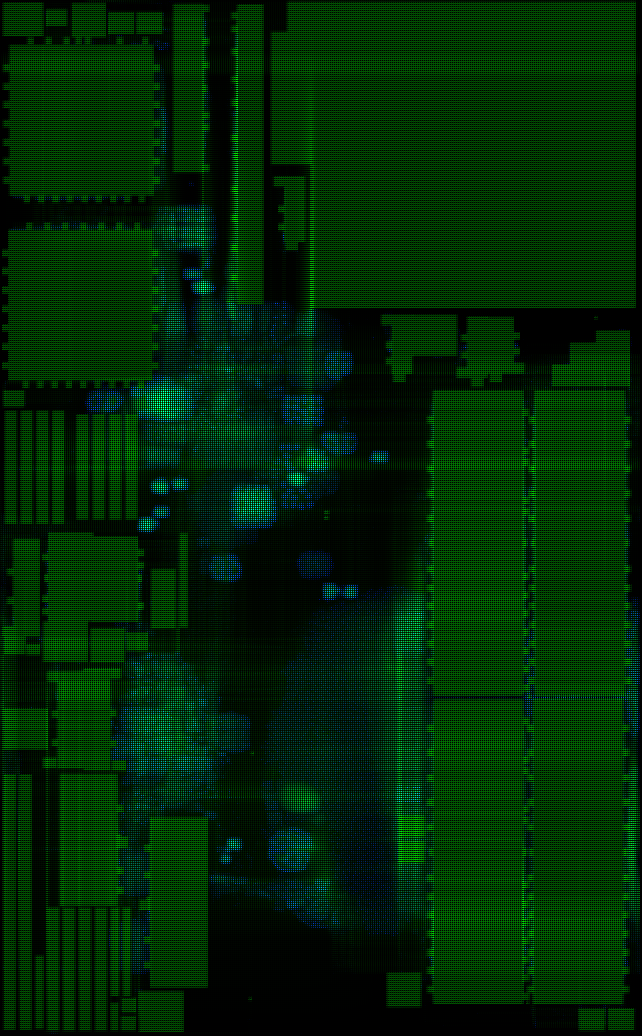}}
	\caption{Input}
	\end{subfigure}
	\qquad
	\begin{subfigure}{0.25\linewidth}
	\centering
	\includegraphics[origin=c,angle=90,width=\linewidth]{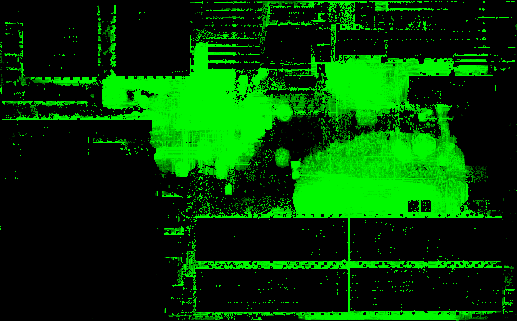}
	\caption{Ground-truth}
	\end{subfigure}
	\qquad
	\begin{subfigure}{0.25\linewidth}
		\centering
		\includegraphics[origin=c,angle=90,width=\linewidth]{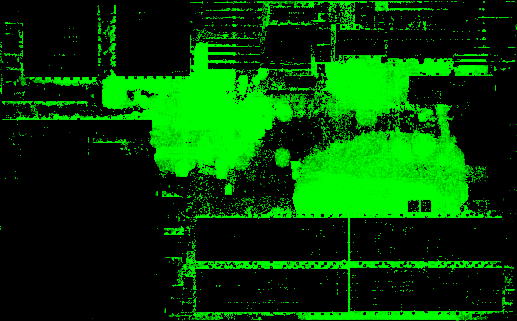}
		\caption{Prediction of {\ours}}
	\end{subfigure}
	\caption{Visualization of congestion maps of circuit \texttt{DAC2012/superblue19} produced by {\ours} versus the ground-truth.}
	\label{fig:visualization}
\end{figure*}

For the placement stage, {\ours} is compared with the following state-of-the-art congestion prediction techniques:
{\ptp} \cite{zhou2019congestion}, {\lhnn} \cite{wang2022lhnn} and {\cgnn} with geometry-sensitive features \cite{yang2022versatile}.
\begin{enumerate}
	\item {\ptp} \cite{zhou2019congestion}, A universal approach that cut a circuit into small rectangles i.e. grids and convert it into RGB channels, where
	the grids are treated as pixels and use image translation methods to output new images, indirectly solving the congestion prediction task.
	\item {\lhnn} \cite{wang2022lhnn} converts the circuits into lattice networks \cite{you2017deep} instead of images, where each grid serves as an internal node in the network and each net, as an external node, is connected to the grids it covers geometrically.
	\item {\cgnn} \cite{yang2022versatile}, as mentioned above, is a versatile graph neural network that designs a heterogeneous graph based on topological and geometrical information. In the placement stage, both the topology and geometry of the circuits are used as the features.
\end{enumerate}

\subsection{Experimental results}
As depicted in Table \ref{tab:logic_ispd} and Table \ref{tab:logic_dac}, the proposed congestion prediction method is first evaluated on circuits during the logic synthesis stage when geometric information is not available. Subsequently, the performance of the method is assessed during the placement stage, as illustrated in Table \ref{tab:place_ispd} and Table \ref{tab:place_dac}. It is important to note that the geometrical methods {\ptp} and {\lhnn} are not designed to account for cells in circuit design, and therefore, their performance is not evaluated on the cell-level. The {\gat} (w. geom.) represents the regular {\gat} model with cell positions incorporated as additional features.

The results show that:
\begin{enumerate}
	\item During the logic synthesis stage, the proposed method, which relies solely on topology information, demonstrates superior performance compared to the majority of other methods. While it does not outperform the top-performing method in terms of the Pearson correlation metric for the \texttt{ISPD2011} dataset at the grid-level and the \texttt{DAC2012} dataset at the cell-level, its performance remains highly competitive.
	\item In the placement stage, the proposed method, which leverages both topology and geometry information, outperforms almost all other methods, with the exception of the Pearson correlation metric for the \texttt{ISPD2011} dataset at the grid-level.
\end{enumerate}

The robust performance of our proposed method, which integrates both topological and geometrical information, is evidenced across various stages of the design process. Figure \ref{fig:visualization} visually demonstrates the effectiveness of our algorithm in predicting congestion, emphasizing its potential as a solid solution for congestion prediction tasks in EDA.

\begin{table}[t]
	\caption{Inference time (second/epoch) results in logical synthesis stage.}
	\label{tab:time_logic}
	\centering
	\normalsize
	\begin{tabular}{@{}ccc@{}}
	\toprule
	Baseline				 & ISPD2011    & DAC2012    \\ \midrule
	{\gcn}         & 10.85      & 10.55      \\
	{\sage}        & 11.60       & 12.00      \\
	{\gat}         & 12.24       & 11.28      \\
	{\congnet}     & 11.58       & 11.36      \\
	{\mpnn}        & 36.17       & 30.44      \\
	{\cgnn}        & 22.78       & 18.51      \\
	{\ours}(w/o. geom.)     & 10.21       & 10.41      \\ \bottomrule
	\end{tabular}
\end{table}

\begin{table}[t]
	\caption{Inference time (second/epoch) results in placement stage.}
	\label{tab:time_place}
	\centering
	\normalsize
	\begin{tabular}{@{}ccc@{}}
	\toprule
		Baseline					& ISPD2011 & DAC2012 \\ \midrule
	{\gat} (w. geom.)      & 11.93    & 11.22    \\
	{\ptp}                 & 1.40     & 1.33    \\
	{\lhnn}               & 63.15    & 102.41   \\
	{\cgnn} (w/o. topo.)  & 14.93    & 14.03   \\
	{\cgnn}               & 15.37    & 16.04   \\
	{\ours}               & 11.48     & 10.81   \\ \bottomrule
	\end{tabular}
\end{table}

\subsection{Runtime comparison}
Table \ref{tab:time_logic} and Table \ref{tab:time_place} report the inference time of {\ours} and other congestion prediction approaches in the logic synthesis stage and the placement stage, respectively. The runtime of all methods are measured on a single NVIDIA RTX-3090 GPU and a AMD Ryzen 9 5950X CPU. In the logical synthesis stage, {\ours} outperforms all the other methods in terms of inference time. In the placement stage, {\ours} demonstrates the best performance against almost all methods except for the {\ptp} which is an image-based method, inherently capable of leveraging GPU acceleration technologies for quicker computations. This experimental results shows that our method also possesses certain advantages in terms of runtime.

\section{Conclusion}

In conclusion, we have introduced a novel approach for congestion prediction in IC design, focusing on the underexplored aspect of spatial label-correlation between different grid cells. Our method, VAriational Label-Correlation Enhancement for congestion prediction (\ours), leverages a regularized risk estimator and a variation inference technique to estimate the label-correlation weight, representing the degree to which each surrounding grid influences the congestion value of the center grid with the consideration that a grid with a higher congestion value tends to suggest higher congestion values in its neighboring grids, revealing an intuitive geometrical relationship worth capturing in a predictive model. The label-correlation thus modeled provides valuable information to refine the training process of the regression model used for congestion prediction. Experimental results on publicly available \texttt{ISPD2011} and \texttt{DAC2012} benchmarks using the superblue circuit designs have demonstrated the superior effectiveness of our method.





\bibliographystyle{plain}
\bibliography{eda_ref}


\end{document}